\documentclass{article}
\usepackage{ijcai23}

\usepackage{times}
\usepackage{soul}
\usepackage{url}
\usepackage[hidelinks]{hyperref}
\usepackage[utf8]{inputenc}
\usepackage[small]{caption}
\usepackage{graphicx}
\usepackage{amsmath}
\usepackage{amsthm}
\usepackage{booktabs}
\usepackage{algorithm}
\usepackage{algorithmic}
\usepackage[switch]{lineno}

\usepackage{epstopdf}
\usepackage{mathtools}
\usepackage{amsfonts}
\usepackage{dsfont}
\usepackage{multirow}
\usepackage{makecell}

\newcommand{\ie}{\textit{i.e.}}

\newcommand{\our}{\mbox{SmartBERT}}
\title{SmartBERT: A Promotion of Dynamic Early Exiting Mechanism for Accelerating BERT Inference}

\author{
Boren Hu\footnotemark[1]\and
Yun Zhu\footnotemark[1]
\and
Jiacheng Li\and
Siliang Tang\footnotemark[2]
\affiliations
Zhejiang University\\
\emails
\{boren,zhuyun\_dcd,lijiacheng,siliang\}@zju.edu.cn
}

\begin{document}

\maketitle

\renewcommand{\thefootnote}{\fnsymbol{footnote}}
\footnotetext[1]{Equal Contribution}
\renewcommand{\thefootnote}{\fnsymbol{footnote}}
\footnotetext[2]{Corresponding Author}

\begin{abstract}
Dynamic early exiting has been proven to improve the inference speed of the pre-trained language model like BERT. However, all samples must go through all consecutive layers before early exiting and more complex samples usually go through more layers, which still exists redundant computation. In this paper, we propose a novel dynamic early exiting combined with layer skipping for BERT inference named \our, which adds a skipping gate and an exiting operator into each layer of BERT. \our~can adaptively skip some layers and adaptively choose whether to exit. Besides, we propose cross-layer contrastive learning and combine it into our training phases to boost the intermediate layers and classifiers which would be beneficial for early exiting. To keep the consistent usage of skipping gates between training and inference phases, we propose a hard weight mechanism during training phase. We conduct experiments on eight classification datasets of the GLUE benchmark. Experimental results show that \our \space achieves 2-3× computation reduction with minimal accuracy drops compared with BERT and our method outperforms previous methods in both efficiency and accuracy. Moreover, in some complex datasets like RTE and WNLI, we prove that the early exiting based on entropy hardly works, and the skipping mechanism is essential for reducing computation. 
\end{abstract}

\section{Introduction}
In recent years, large-scale pre-trained language models(PLMs) such as BERT \cite{devlin-etal-2019-bert}, GPT \cite{radford2018improving}, ALBERT \cite{lan2019albert}, and XLNET \cite{yang2019xlnet} RoBERTa \cite{liu2019roberta} have made significant progress in the field of natural language processing(NLP). However, these models usually require large computation resources. It is difficult to deploy these models in the case of limited resources. To reduce redundant computation, various approaches have been proposed, including network pruning \cite{li2016pruning,he2017channel}, weight quantization \cite{jacob2018quantization}, knowledge distillation \cite{hinton2015distilling,sanh2019distilbert,jiao2019tinybert} and dynamic early exiting \cite{xin2020deebert,liu2020fastbert,schwartz2020right}.

In this work, we mainly focus on dynamic early exiting methods. Such methods do not change the original network structure, and they only add some light plugins in the original network to decide whether early exit or not in each layer which can reduce plenty of computation while keeping comparable performance.

Although current dynamic early exiting technology has shown excellent characteristics, samples must go through all consecutive layers before early exiting, and complex samples have to go through nearly all layers. That is, complex samples have more redundant computation. In the field of computer vision, Highway Network \cite{srivastava2015highway} and SkipNet \cite{wang2018skipnet} have proved that redundant computations exist in neural networks and some blocks~(layers) can be skipped directly. Motivated by these works, we assume that some layers also can be skipped directly before early exiting for BERT to further reduce computation. 

In this paper, firstly, we introduce a novel dynamic early exiting combined with layer skipping technique for BERT inference named \our, which plugs a skipping gate and an exiting operator into each layer of BERT and performs adaptive inference based on the principle of the higher priority of skipping than exiting. We use the widely used PLM BERT~\cite{devlin-etal-2019-bert} as the backbone, and our method can be extended to other PLMs. 

Secondly, in order to address the inconsistent usage of skipping gates between the train and inference stage, we design a hard weight mechanism. For adopting the transformation from soft to hard, we proffer a special training way.
Thirdly, we propose cross-layer contrastive learning and combine it into our training phases to boost the intermediate layers and classifiers to achieve more efficient computation.
Moreover, We conducted experiments on eight classification datasets of the GLUE benchmark\cite{wang2018glue}, which shows that \our \space achieves 2-3× computation reduction with minimal accuracy drops compared with BERT and has a better performance than other dynamic early exiting models at the same computational cost. We also prove skipping mechanism is highly effective in some complex datasets like RTE and WNLI.

Our main contributions can be summarized
as follows: 
\begin{itemize}
\item We propose a novel dynamic early exiting method combined with layer skipping for BERT inference. And we design a hard weight mechanism and a special training way for consistent usage and better training respectively.
\item We propose cross-layer contrastive learning into the training phase to boost the intermediate layers and classifiers, which is beneficial for early exiting.
\item Our method can achieve 2-3x computation reduction with comparable performance and outperforms previous methods in both efficiency and accuracy. In some complex datasets like RTE and WNLI, our method can save the quantity of computation time (2-4x) compared to current early existing methods while achieving better performance. 
\end{itemize}

\section{Related work}
\subsection{Model Compression}
Many pre-trained language models~(PLMs) have emerged in the last few years. Although these PLMs have achieved high scores in many NLP tasks, their inference time is slow, and the cost of calculation is expensive. One of the most representative models is BERT \cite{devlin-etal-2019-bert}, which has made remarkable improvements in many NLP tasks. However, the inference speed of BERT is criticized. Therefore, a series of methods of model compression have been proposed to solve the above problems. Knowledge distillation \cite{hinton2015distilling} aims to transfer knowledge from the teacher model to the student model, which is applied in DistillBERT \cite{sanh2019distilbert} and TinyBERT \cite{jiao2019tinybert}. ALBERT \cite{lan2019albert} uses sharing parameters to greatly reduce the number of parameters and memory consumption. Q8BERT \cite{zafrir2019q8bert} use symmetric linear quantization \cite{jacob2018quantization} to reduce the number of bits about the parameters of BERT. For Pruning, \cite{DBLP:journals/corr/abs-2002-08307} mainly remove the unimportant part based on gradients of weights. Although these methods can improve the inference time of BERT, they can not adaptively change the architecture of the model according to the complexity of each sample. For example, simple samples must go through all layers for these methods. However, these samples may only go through early layers in early exiting models. 

\subsection{Adaptive Inference}
Adaptive inference can adaptively change the architecture of the model according to the complexity of samples. DeeBERT \cite{xin2020deebert} chooses whether to early exit according to the entropy of the output distribution on each layer. FastBERT divides the early exiting classifiers into student classifiers and the teacher classifier, and then trains all student classifiers through self-distillation. FastBERT also uses the early exit based on entropy. FastBERT and DeeBERT can adaptively adjust the model size according to different samples. However, samples need to go through all layers before early exiting, and if they are complex enough, they need to go through the whole model. Hence, we introduce cross-layer contrastive learning to obtain more powerful classifiers and use layer skipping mechanism to reduce redundancy further.

\subsection{Contrastive Learning}
Recently, contrastive learning (CL) has made significant progress in various domains. Especially, unsupervised CL can exploit a bulk of unlabelled data to train a model with generalization which can even surpass the model trained under supervised training in some situations. The aim of CL is to maximize the agreement between positive views which are jointly sampled and disparting negative views in the representation space. Pioneering works were mainly proposed in the CV domain~\cite{he2020momentum,chen2020simple}. Then in the graph domain, there are plentiful follow-up works \cite{you2020graph,zhu2022rosa} which borrow ideas from pioneering works. However, in the field of natural language processing, there are few inventive works \cite{gao2021simcse} that have shown up in recent years. The main difficulty in NLP domain is that augmentation is hard to design. Besides, previous works mainly focus on the quality of the final representations which may neglect the intermediate layers. But in our work, we hope each layer can be competent for good classification to early exit. So, we propose \emph{cross-layer} contrastive learning in our method for both the first stage and the second training stage. In the \emph{cross-layer} contrastive learning, we do not rely on data augmentation and assume that each token should have similar semantics across the consecutive layers because their representations do not change drastically. The same tokens will be treated as positive pairs in the consecutive layers, otherwise they shall be negative pairs.

\section{Methodology}

In this section, we first introduce the architecture of \our, which combines the early exiting mechanism and layer skipping mechanism. Secondly, we will illustrate how to combine cross-layer contrastive learning into the training phase to obtain powerful classifiers. Lastly, we introduce a hard weight mechanism and a special training way for skipping gates, further keeping the consistent usage of skipping gates between the training and inference phases.
\subsection{Preliminaries}
Given an input sentence $\mathbf{S}=\left\{\mathbf{w}_{1}, \mathbf{w}_{2}, \ldots, \mathbf{w}_{N}\right\} \in \mathbb{R}^{N \times W}\ $, $N$ and $L$ stand for the number of tokens and encoder layers respectively. $\mathbf{X}^{i} \in \mathbb{R}^{N \times D}$ represents the output of the $i^{\text{th}}$ encoder layer $\mathcal{E}^{i}(\cdot)$. $\mathbf{X}^{0}$ stands for the output of embedding layer. $\mathcal{G}^{i}(\cdot)$ and $\mathcal{C}^{i}(\cdot)$ represent the skipping gate and classifier, and the superscript means they are inserted in $i^{\text{th}}$ layer.
\subsection{Model Design}
Different to \cite{liu2020fastbert,xin2020deebert} whose each layer will introduce one exiting classifier, we additionally plug a skipping gate into each layer. In order to understand \our \space better, we will first introduce the skipping gates and then introduce exiting classifiers. 

\begin{figure}[ht]
    \centering
    \includegraphics[scale=0.6]{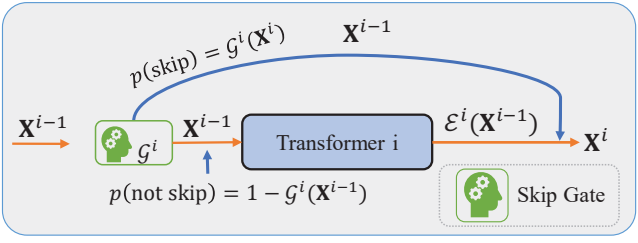}
    \caption{The skipping gate has $p(\text{skip})$ probability to skip the current transformer layer and has $p(\text{not skip})$ probability not to skip. In the inference phase, when $p(\text{skip})$ is greater than $p(\text{not skip})$, the gate will choose to skip, otherwise not to skip.}
    \label{fig:pic2}
\end{figure}

The skipping gate aims to decide whether to execute or bypass the current layer like Figure \ref{fig:pic2}. 

With the skipping gate $\mathcal{G}^{i}(\cdot)$, the output of $i^{th}$ encoder block can be defined as
\begin{equation}
 \mathbf{X}^i = \mathcal{G}^{i}(\mathbf{X}^{i-1})\mathbf{X}^{i-1} + (1 - \mathcal{G}^{i}(\mathbf{X}^{i-1}))\mathcal{E}^{i}(\mathbf{X}^{i-1}), 
\label{eq2}
\end{equation}
where $\mathcal{G}^{i}(\mathbf{X}^{i-1})\subseteq[0,1]$ is the output of the $i^{th}$ skipping gate which represents the probability of skipping the current layer. In the training phase, $\mathcal{G}^{i}(\mathbf{X}^{i-1})$ is a continuous value in the range 0 to 1. However, in the inference phase, $\mathcal{G}^{i}(\mathbf{X}^{i-1})$ is a discrete value that is 0 or 1, and the inconsistent usage of skipping gates between the training and inference phase will be discussed in Section~\ref{hard_weight}. 

Next, we will introduce early exiting classifiers. All early exiting classifiers use the early exiting mechanism based on the entropy, and we define the output of classifiers as

\begin{equation}
 \mathbf{z}^i = \mathcal{C}^{i}(\mathbf{X}^i), 
\label{eq3}
\end{equation}
where $\mathbf{z}^i \in \mathbb{R}^{1 \times C}$ represents the probability of each category.

\subsection{Training}
\our \space is composed of pre-trained BERT, skipping gates, and early exiting classifiers. For stability, we use two stages training strategy for the model, and different stages will train different parts of the model. In this section, we will illustrate two stages separately.

\subsubsection{First Training Stage}
\begin{figure}[ht]
    \centering
    \includegraphics[scale=0.48]{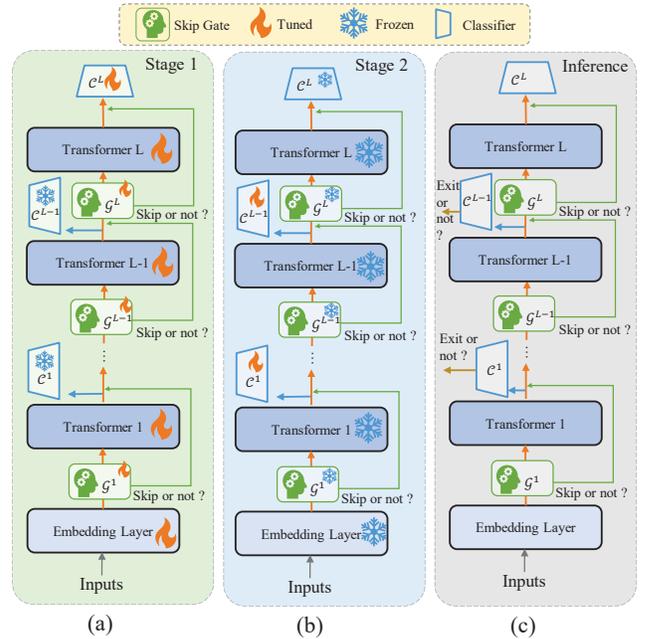}
    \caption{The processes of two-stage training stages and inference phase. (a) and (b) show the first and second stages of training, respectively. Note that different components will be trained in different stages. (c) shows the fast inference process. In each layer, the skipping will decide whether to skip the current layer. If it chooses to skip, then the input directly enters the next gate. Otherwise, the input will be fed into the current layer, and the corresponding classifier will decide whether to exit.}
    \label{fig:pic3}
\end{figure}

In the first stage, we train skipping gates and fine-tune pre-trained BERT in a joint way. Other components will be fixed except the last classifier, as shown in Figure \ref{fig:pic3}a. The total loss of the first stage consists of three main parts: classification error, the sloth of skipping, and cross-layer contrastive loss. It can be formalized as
\begin{equation}
\mathcal{L}_{\text {first}} = \mathcal{L}_{\text{CE}}(\mathbf{z}^L,\mathbf{y}) + \frac{\lambda}{\sum\limits^n_{i=1}{\mathcal{G}^i(\mathbf{X}^i)}} + \eta\mathcal{L}_{\text{contra\_1}},
\label{eq4}
\end{equation}
where the first term represents the classification error(e.g., cross-entropy loss) like Equation \ref{eq3}. The second term indicates the sloth of skipping. In order to encourage the gates to skip to some extent rather than only considering the performance, we introduce the second term as a regular term. To enlarge the output of skipping gates. And $\lambda,\eta \in[0,1]$ are scaling factors that are used to balance the influence of the regular terms on the loss. The third term represents a \emph{cross-layer} contrastive loss, which mainly aims to obtain a more powerful BERT in this stage. 

In cross-layer contrastive learning (CCL), we assume that each token should have similar semantics across the consecutive layers because their representations do not change drastically. The experiments in Appendix C can support our motivation. The same tokens will be considered as positive pairs across consecutive layers. Otherwise, they shall be deemed to be negative pairs. The CCL loss in the first training stage can be formulated as
\begin{equation}
\begin{split}
    \mathcal{L}_{\text{contra\_1}} = \frac{1}{N}\frac{1}{L-1}\sum_{i=1}^{L-1}\sum_{m=1}^{N}
    -\log \frac{e^{\textnormal{s}({\mathbf{p}}^{i}_m, {\mathbf{p}}^{i+1}_m)/ \tau}}{\sum_{k=1}^{N}e^{\textnormal{s}({\mathbf{p}}^{i}_m, {\mathbf{p}}^{i+1}_k)/ \tau} },
\end{split}
\label{contra1}
\end{equation}
where $\mathbf{p}_m^i=f(\boldsymbol{x}_m^i)$, $\boldsymbol{x}_m^i$ represents the $m^{\text{th}}$ word representation in the $i^{\text{th}}$ layer. $f(\cdot)$ is a projection function that maps representations to another latent space where the contrastive loss is calculated. $\textnormal{s}(x,y)$ represents a score function (e.g., cosine similarity), and $\tau$ is temperature parameter. In this training stage, we only boost the transformer layer by the cross-layer contrastive loss.

\subsubsection{Second Training Stage} \label{second}
\begin{figure}[ht]
    \centering
    \includegraphics[scale=0.6]{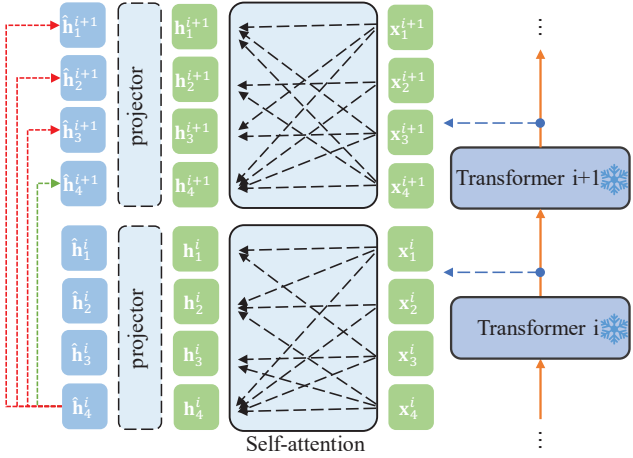}
    \caption{The pipeline of cross-layer contrastive learning in the second stage. 
    The green line~(\ie, the same token in the consecutive layers) indicates positive pairs, and red lines~(\ie, otherwise) represent negative pairs.}
    \label{fig:contra}
\end{figure}

In the second stage, the trained parameters of the first stage will be frozen, and all classifiers except the last one will be trained as shown in Figure~\ref{fig:pic3}b. In order to obtain more powerful classifiers, we also adopt CCL into the second training phase.
Specifically, the token representations obtained by frozen transformer layers will be fed into classifiers firstly. Secondly, we extract the hidden states that only go through the first layer~(\ie, self-attention layer) of classifiers, and these hidden states will be fed into the projection head to map them into the space where contrastive loss is occupied. Lastly, the contrastive loss~(\ie, InfoNCE loss) is employed. Figure~\ref{fig:contra} shows the overall pipeline. The CCL loss in the second training stage can be formalized as
\begin{equation}
\begin{split}
    \ell(\hat{\mathbf{h}}^{i}_m, \hat{\mathbf{h}}^{i+1}_m) = 
    -\log \frac{e^{\textnormal{s}(\hat{\mathbf{h}}^{i}_m, \hat{\mathbf{h}}^{i+1}_m)/ \tau}}{\sum_{k=1}^{N}e^{\textnormal{s}(\hat{\mathbf{h}}^{i}_m, \hat{\mathbf{h}}^{i+1}_k)/ \tau} },
\end{split}
\label{contra}
\end{equation}
where $\hat{\mathbf{h}}^{i}_m=f(\mathbf{h}^{i}_m), \mathbf{h}^{i}_m=\text{Self-Attention}(\mathbf{X}^i)[m]$, $\text{Self-Attention}(\cdot)$ represents a self-attention function and $[m]$ means we only select $m$-th token. Other notations have the same meanings as in Equation~\ref{contra1}.
Enumerating all tokens in all layers, the overall loss is:

\begin{equation}
    \mathcal{L}_{\text{contra\_2}}=\frac{1}{N}\frac{1}{L-1}\sum_{i=1}^{L-1}\sum_{m=1}^{N}\ell(\hat{\mathbf{h}}^{i}_m, \hat{\mathbf{h}}^{i+1}_m).
\end{equation}
In this way, the exiting classifiers can help each other in the consecutive layers, which will assist our model early exiting. 
In detail, our classifier is similar to \cite{liu2020fastbert}, which consists of one attention layer and one linear layer. Each token in the $i$-th layer will acquire its high-level representation $\mathbf{H}^i$ through the attention layer. Through Equation~\ref{contra}, the self-attention layer will be boosted so that we can empower the classifier to produce more confident results which are helpful for early exiting based on entropy.

For keeping consistent with the inference phase, \our \space uses the output of skipping gates to update the hidden states of the current layer like Equation \ref{eq2}. Next, we define the loss of each classifier as
\begin{equation}
\mathcal{L}_{\text{CE}}^{i} = \mathcal{L}_{\text{CE}}(\mathbf{z}^i,\mathbf{y}).
\label{eq5}
\end{equation}
In the second stage, the loss function is the sum of the cross-entropy loss of classifiers with cross-layer contrastive loss:
\begin{equation}
\mathcal{L}_{\text{second}} = \sum\limits^{L-1}_{i=1}\mathcal{L}_{\text{CE}}^{i}+\eta\mathcal{L}_{\text{contra\_2}}.
\label{eq6}
\end{equation}

\subsection{Inference}
After finishing the two-stage training, \our\space can take advantage of skipping gates and early exiting to accelerate inference considering the complexity of samples. 
Only when the output of skipping gates is greater than or equal to 0.5, the model will decide to skip. The specific formula follows:

\begin{equation}
\mathbf{X}^i = 
\left\{
             \begin{array}{lr}
             \mathbf{X}^{i-1}&,\quad \text{if}~ \mathcal{G}^{i}(\mathbf{X}^{i-1}) \geq 0.5 \\
             \\
             \mathcal{E}^i(\mathbf{X}^{i-1})&,\quad \text{if}~ \mathcal{G}^{i}(\mathbf{X}^{i-1}) < 0.5
             \end{array}
\label{eq7}
\right.
\end{equation}
When skipping gates choose not to skip, the current early exiting classifier will be used, and all classifiers use an early exiting mechanism based on the entropy:
\begin{equation}
\left\{
             \begin{array}{cl}
             \text{Exit}&,\quad \text{if}~\operatorname{Entropy}(\mathbf{z}^i) < S \\
             \\
             \text{Continue}&,\quad \text{otherwise}
             \end{array}
\label{eq8}
\right.
\end{equation}
where $\mathbf{z}^i$ represents a categorical distribution of the $i^{th}$ classifier, and $S$ is the entropy threshold for early exiting, which is set manually. The entire inference algorithm is in Appendix A.

\subsection{Hard Weight Mechanism} \label{hard_weight}
\label{sec:3.4}
In the previous sections, we have introduced \our \space training and inference. However, we find the inconsistent usage of skipping gates between the training and inference phase. Specifically, in the training phase, the output of skipping gates is a continuous probability value between 0 and 1 like Equation~\ref{eq2}, and in the inference phase, we will discretize the output of skipping gates like Equation \ref{eq7} for acceleration. To reduce the impact of inconsistent usage of skipping gates, we introduce a hard-weight mechanism for skipping gates in the training phase. And the hard weight mechanism is differentiable, we use the trick:
\begin{equation}
G_{\text{hard}} = \mathds{1}_{G_{\text{soft}\geq 0.5}} - \Tilde{G}_{\text{soft}} + G_{\text{soft}},
\end{equation}
where $\mathds{1}_{G_{\text{soft}\geq 0.5}}$ is an indicator function which returns 1 if $G_{\text{soft}\geq 0.5}$ otherwise 0, and $\Tilde{G}_{\text{soft}}$ is the result of truncating the gradient of  $G_{\text{soft}}$. With this trick, we achieve two things: (1) makes the output value exactly one-hot (since we add then subtract soft value), (2) makes the gradient equal to the soft gradient (since we strip all other gradients). In the training phase, we define the output of skipping gates as the soft weight $G_{\textrm{soft}}$. To keep the consistent usage of skipping gates between training and inference phases, we turn the soft weight $G_{\textrm{soft}}$ into discrete values~(0 or 1), and keep the gradient equal to Equation \ref{eq9}. In this way, the output of the skipping gate becomes a discrete skipping decision in the training phase, which is consistent with the inference phase:
\begin{equation}
G_{\textrm{hard}} = 
\left\{
             \begin{array}{cl}
             1&, \quad \text{fp}, G_{\textrm{soft}}\geq 0.5  \\
             0&,\quad \text{fp}, G_{\textrm{soft}}< 0.5 \\
             G_{\textrm{soft}}&,\quad \text{backward pass}
             \end{array}
\label{eq9}
\right.
\end{equation}
where fp represents forward pass, $G_{\textrm{soft}}$ is the soft weight and $G_{\textrm{hard}}$ is the hard weight.
\subsection{A Special Training Way for Skipping Gates}
Although the hard weight mechanism can reduce the impact of inconsistent usage of skipping gates, directly using the hard weight mechanism  will lead to some layers can not be well fine-tuned because they may be skipped. However, these layers may be used in the inference phase, which will result in unsatisfactory performance. To ensure all layers can be trained and keep the consistent usage of skipping gates in the training and inference phase, we introduce a special training way from soft weight mechanism to hard weight mechanism for skipping gates.

We use the transition from the soft weight mechanism to the hard weight mechanism to address the above problems. First, we use the soft weight mechanism to warm up all skipping gates and transformer layers and then use the hard weight mechanism to train them. Moreover, the soft weight mechanism aims to fine-tune these layers, and the hard weight mechanism is used for keeping the consistent usage of skipping gates between training and inference phases.

\begin{table*}[th]
\scalebox{0.68}{
\begin{tabular}{c|cc|cc|cc|cc|cc|cc|cc|cc|cc}
\toprule
\makecell{Dataset/\\ Model}      & \multicolumn{2}{c|}{MRPC}  & \multicolumn{2}{c|}{SST-2}  & \multicolumn{2}{c|}{QNLI}     
                      & \multicolumn{2}{c|}{QQP}   & \multicolumn{2}{c|}{MNLI}   & \multicolumn{2}{c|}{MNLI-mm}       
                      & \multicolumn{2}{c|}{CoLA}  & \multicolumn{2}{c|}{RTE}    & \multicolumn{2}{c}{WNLI}    
\\ \midrule
& F1            & \begin{tabular}[c]{@{}c@{}}FLOPs\\ (cost)\end{tabular}           
& Acc           & \begin{tabular}[c]{@{}c@{}}FLOPs\\ (cost)\end{tabular}           
& Acc           & \begin{tabular}[c]{@{}c@{}}FLOPs\\ (cost)\end{tabular}           
& F1            & \begin{tabular}[c]{@{}c@{}}FLOPs\\ (cost)\end{tabular}           
& Acc           & \begin{tabular}[c]{@{}c@{}}FLOPs\\ (cost)\end{tabular}           
& Acc           & \begin{tabular}[c]{@{}c@{}}FLOPs\\ (cost)\end{tabular}           
& Mcc           & \begin{tabular}[c]{@{}c@{}}FLOPs\\ (cost)\end{tabular}           
& Acc           & \begin{tabular}[c]{@{}c@{}}FLOPs\\ (cost)\end{tabular}           
& Acc           & \begin{tabular}[c]{@{}c@{}}FLOPs\\ (cost)\end{tabular}           
\\ \midrule
BERT                                                       & 88.5          & \begin{tabular}[c]{@{}c@{}}21744M\\ (100\%)\end{tabular}        & \textbf{93.2}
& \begin{tabular}[c]{@{}c@{}}21744M\\ (100\%)\end{tabular}        & \textbf{88.1} & \begin{tabular}[c]{@{}c@{}}21744M\\ (100\%)\end{tabular}        & \textbf{84.4} & \begin{tabular}[c]{@{}c@{}}21744M\\ (100\%)\end{tabular}        & \textbf{84.6} & \begin{tabular}[c]{@{}c@{}}21744M\\ (100\%)\end{tabular}        & \textbf{84.8} & \begin{tabular}[c]{@{}c@{}}21744M\\ (100\%)\end{tabular}        & 50.8          & \begin{tabular}[c]{@{}c@{}}21744M\\ (100\%)\end{tabular}        & \textbf{61.0} & \begin{tabular}[c]{@{}c@{}}21744M\\ (100\%)\end{tabular}        & 56.3 & \begin{tabular}[c]{@{}c@{}}21744M\\ (100\%)\end{tabular}        
\\ \midrule
DistillBERT                                                       & 87.0          & \textbf{\begin{tabular}[c]{@{}c@{}}10872M\\ (50\%)\end{tabular} }       & 91.5 & \begin{tabular}[c]{@{}c@{}}10872M\\ (50\%)\end{tabular}        & 86.8 & \begin{tabular}[c]{@{}c@{}}10872M\\ (50\%)\end{tabular}        & 82.4 & \begin{tabular}[c]{@{}c@{}}10872M\\ (50\%)\end{tabular}        & 81.9 & \begin{tabular}[c]{@{}c@{}}10872M\\ (50\%)\end{tabular}        & 81.8 & \begin{tabular}[c]{@{}c@{}}10872M\\ (50\%)\end{tabular}        & 47.5          & \begin{tabular}[c]{@{}c@{}}10872M\\ (50\%)\end{tabular}        & 59.2 & \textbf{\begin{tabular}[c]{@{}c@{}}10872M\\ (50\%)\end{tabular} }       & 55.3 & \begin{tabular}[c]{@{}c@{}}10872M\\ (50\%)\end{tabular}        
\\ \midrule
\begin{tabular}[c]{@{}c@{}}DeeBERT\\ (S=0.1)\end{tabular}  
& 88.4          & \begin{tabular}[c]{@{}c@{}}21744M\\ (100\%)\end{tabular}        
& 92.4          & \begin{tabular}[c]{@{}c@{}}13900M\\ (63\%)\end{tabular}          
& 87.5          & \begin{tabular}[c]{@{}c@{}}20008M\\ (92\%)\end{tabular}         
& 84.2          & \begin{tabular}[c]{@{}c@{}}17578M\\ (80\%)\end{tabular}          
& 84.2          & \begin{tabular}[c]{@{}c@{}}20796M\\ (95\%)\end{tabular}         
& 84.3          & \begin{tabular}[c]{@{}c@{}}20786M\\ (95\%)\end{tabular}         
& 49.9          & \begin{tabular}[c]{@{}c@{}}21840M\\ (100\%)\end{tabular}         
& 59.9          & \begin{tabular}[c]{@{}c@{}}21758M\\ (100\%)\end{tabular}       
& 56.3          & \begin{tabular}[c]{@{}c@{}}21758M\\ (100\%)\end{tabular}        \\
\begin{tabular}[c]{@{}c@{}}DeeBERT\\ (S=0.3)\end{tabular}  
& 88.2          & \begin{tabular}[c]{@{}c@{}}19726M\\ (90\%)\end{tabular}          
& 90.3          & \begin{tabular}[c]{@{}c@{}}10874M\\ (50\%)\end{tabular}          
& 87.1          & \begin{tabular}[c]{@{}c@{}}15770M\\ (72\%)\end{tabular}          
& 83.1          & \begin{tabular}[c]{@{}c@{}}14750M\\ (67\%)\end{tabular}          
& 84.2          & \begin{tabular}[c]{@{}c@{}}19698M\\ (90\%)\end{tabular}          
& 84.1          & \begin{tabular}[c]{@{}c@{}}18148M\\ (83\%)\end{tabular}          
& 49.9          & \begin{tabular}[c]{@{}c@{}}16850M\\ (77\%)\end{tabular}         
& 59.9          & \begin{tabular}[c]{@{}c@{}}21718M\\ (99\%)\end{tabular}         
& 56.3          & \begin{tabular}[c]{@{}c@{}}21758M\\ (100\%)\end{tabular}        \\
\begin{tabular}[c]{@{}c@{}}DeeBERT\\ (S=0.5)\end{tabular}  
& 87.2          & \begin{tabular}[c]{@{}c@{}}13540M\\ (62\%)\end{tabular} 
& 85.2          & \begin{tabular}[c]{@{}c@{}}7468M\\ (34\%)\end{tabular}          
& 84.9          & \begin{tabular}[c]{@{}c@{}}10502M\\ (48\%)\end{tabular}          
& 74.5          & \begin{tabular}[c]{@{}c@{}}8744M\\ (40\%)\end{tabular}          
& 83.9          & \begin{tabular}[c]{@{}c@{}}18728M\\ (86\%)\end{tabular}          
& 83.1          & \begin{tabular}[c]{@{}c@{}}16362M\\ (75\%)\end{tabular}         
& 49.9          & \begin{tabular}[c]{@{}c@{}}15130M\\ (70\%)\end{tabular}         
& 59.9          & \begin{tabular}[c]{@{}c@{}}21332M\\ (98\%)\end{tabular}         
& 56.3          & \begin{tabular}[c]{@{}c@{}}21758M\\ (100\%)\end{tabular}        
\\ \midrule
\begin{tabular}[c]{@{}c@{}}FastBERT\\ (S=0.1)\end{tabular} 
& 88.5          & \begin{tabular}[c]{@{}c@{}}22196M\\ (102\%)\end{tabular}        
& 92.4          & \begin{tabular}[c]{@{}c@{}}11094M\\ (51\%)\end{tabular}          
& 87.5          & \begin{tabular}[c]{@{}c@{}}19684M\\ (90\%)\end{tabular}          
& 83.9          & \begin{tabular}[c]{@{}c@{}}15808M\\ (72\%)\end{tabular}          
& 83.4          & \begin{tabular}[c]{@{}c@{}}19980M\\ (91\%)\end{tabular}         
& 84.0          & \begin{tabular}[c]{@{}c@{}}19116M\\ (87\%)\end{tabular}       
& 49.8          & \begin{tabular}[c]{@{}c@{}}20846M\\ (95\%)\end{tabular}        
& 59.4          & \begin{tabular}[c]{@{}c@{}}22196M\\ (102\%)\end{tabular}        
& 56.3          & \begin{tabular}[c]{@{}c@{}}22196M\\ (102\%)\end{tabular}        \\
\begin{tabular}[c]{@{}c@{}}FastBERT\\ (S=0.3)\end{tabular} 
& 88.5          & \begin{tabular}[c]{@{}c@{}}20278M\\ (93\%)\end{tabular}         
& 90.7          & \begin{tabular}[c]{@{}c@{}}6704M\\ (30\%)\end{tabular}          
& 86.8          & \begin{tabular}[c]{@{}c@{}}13096M\\ (60\%)\end{tabular}          
& 82.9          & \begin{tabular}[c]{@{}c@{}}10586M\\ (48\%)\end{tabular}          
& 83.1          & \begin{tabular}[c]{@{}c@{}}15468M\\ (71\%)\end{tabular}          
& 83.6          & \begin{tabular}[c]{@{}c@{}}15358M\\ (70\%)\end{tabular}         
& 49.5          & \begin{tabular}[c]{@{}c@{}}15262M\\ (70\%)\end{tabular}         
& 59.4          & \begin{tabular}[c]{@{}c@{}}22196M\\ (102\%)\end{tabular}       
& 56.3          & \begin{tabular}[c]{@{}c@{}}22196M\\ (102\%)\end{tabular}        \\
\begin{tabular}[c]{@{}c@{}}FastBERT\\ (S=0.5)\end{tabular} 
& 88.3          & \begin{tabular}[c]{@{}c@{}}16478M\\ (75\%)\end{tabular}         
& 86.8          & \begin{tabular}[c]{@{}c@{}}4076M\\ (18\%)\end{tabular}          
& 84.5          & \begin{tabular}[c]{@{}c@{}}7964M\\ (36\%)\end{tabular}          
& 78.2          & \begin{tabular}[c]{@{}c@{}}6064M\\ (27\%)\end{tabular}          
& 81.3          & \begin{tabular}[c]{@{}c@{}}11958M\\ (54\%)\end{tabular}          
& 81.3          & \begin{tabular}[c]{@{}c@{}}11020M\\ (50\%)\end{tabular}         
& 44.4          & \begin{tabular}[c]{@{}c@{}}12700M\\ (58\%)\end{tabular}         
& 59.4          & \begin{tabular}[c]{@{}c@{}}22196M\\ (102\%)\end{tabular}        
& 56.3          & \begin{tabular}[c]{@{}c@{}}22196M\\ (102\%)\end{tabular}        
\\ \midrule
\begin{tabular}[c]{@{}c@{}}Ours\\ (S=0.1)\end{tabular}     
& \textbf{89.8} & \begin{tabular}[c]{@{}c@{}}21390M\\ (98\%)\end{tabular}         
& 93.1          & \begin{tabular}[c]{@{}c@{}}9040M\\ (41\%)\end{tabular}          
& 87.9          & \begin{tabular}[c]{@{}c@{}}16512M\\ (75\%)\end{tabular}          
& 84.1          & \begin{tabular}[c]{@{}c@{}}13128M\\ (60\%)\end{tabular}          
& 84.4          & \begin{tabular}[c]{@{}c@{}}18436M\\ (84\%)\end{tabular}          
& 84.6          & \begin{tabular}[c]{@{}c@{}}18196M\\ (83\%)\end{tabular}          
& \textbf{51.5} & \begin{tabular}[c]{@{}c@{}}20568M\\ (94\%)\end{tabular}         
& 60.6          & \begin{tabular}[c]{@{}c@{}}12684M\\ (58\%)\end{tabular} 
& 56.3  & \textbf{\begin{tabular}[c]{@{}c@{}}6768M\\ (31\%)\end{tabular}} \\
\begin{tabular}[c]{@{}c@{}}Ours\\ (S=0.3)\end{tabular}     
& 89.6          & \begin{tabular}[c]{@{}c@{}}17884M\\ (82\%)\end{tabular}         
& 91.9          & \begin{tabular}[c]{@{}c@{}}5706M\\ (26\%)\end{tabular}          
& 87.0          & \begin{tabular}[c]{@{}c@{}}10878M\\ (50\%)\end{tabular}          
& 82.5          & \begin{tabular}[c]{@{}c@{}}8610M\\ (39\%)\end{tabular}          
& 83.6          & \begin{tabular}[c]{@{}c@{}}14442M\\ (66\%)\end{tabular}          
& 83.9          & \begin{tabular}[c]{@{}c@{}}13936M\\ (64\%)\end{tabular}          
& 51.3          & \begin{tabular}[c]{@{}c@{}}15072M\\ (69\%)\end{tabular}         
& 60.6          & \begin{tabular}[c]{@{}c@{}}12684M\\ (58\%)\end{tabular} 
& 56.3 & \textbf{\begin{tabular}[c]{@{}c@{}}6768M\\ (31\%)\end{tabular}} \\
\begin{tabular}[c]{@{}c@{}}Ours\\ (S=0.5)\end{tabular}     
& 88.9          & \begin{tabular}[c]{@{}c@{}}14348M\\ (65\%)\end{tabular}          
& 87.7          & \textbf{\begin{tabular}[c]{@{}c@{}}3608M\\ (16\%)\end{tabular}}
& 84.9          & \textbf{\begin{tabular}[c]{@{}c@{}}7020M\\ (32\%)\end{tabular}}
& 78.0          & \textbf{\begin{tabular}[c]{@{}c@{}}5130M\\ (23\%)\end{tabular}} 
& 82.4          & \textbf{\begin{tabular}[c]{@{}c@{}}10450M\\ (49\%)\end{tabular}}
& 82.0          & \textbf{\begin{tabular}[c]{@{}c@{}}10570M\\ (48\%)\end{tabular}}
& 49.9          & \textbf{\begin{tabular}[c]{@{}c@{}}10468M\\ (49\%)\end{tabular}}
& 60.6          & \begin{tabular}[c]{@{}c@{}}12684M\\ (58\%)\end{tabular} 
& 56.3 & \textbf{\begin{tabular}[c]{@{}c@{}}6768M\\ (31\%)\end{tabular}} \\ 
\bottomrule
\end{tabular}}
\caption{Comparison between baselines(BERT,DistillBERT,FastBERT,DeeBERT) and \our \space on the GLUE benchmark. \emph{FLOPs} are multiply–accumulate operations which represent computational complexity. \emph{S} represents the entropy threshold, and \emph{cost} is the computational cost.}
\label{tab:tabel2}
\end{table*}

\begin{figure*}[htpb]
    \centering
    \includegraphics[scale=0.65]{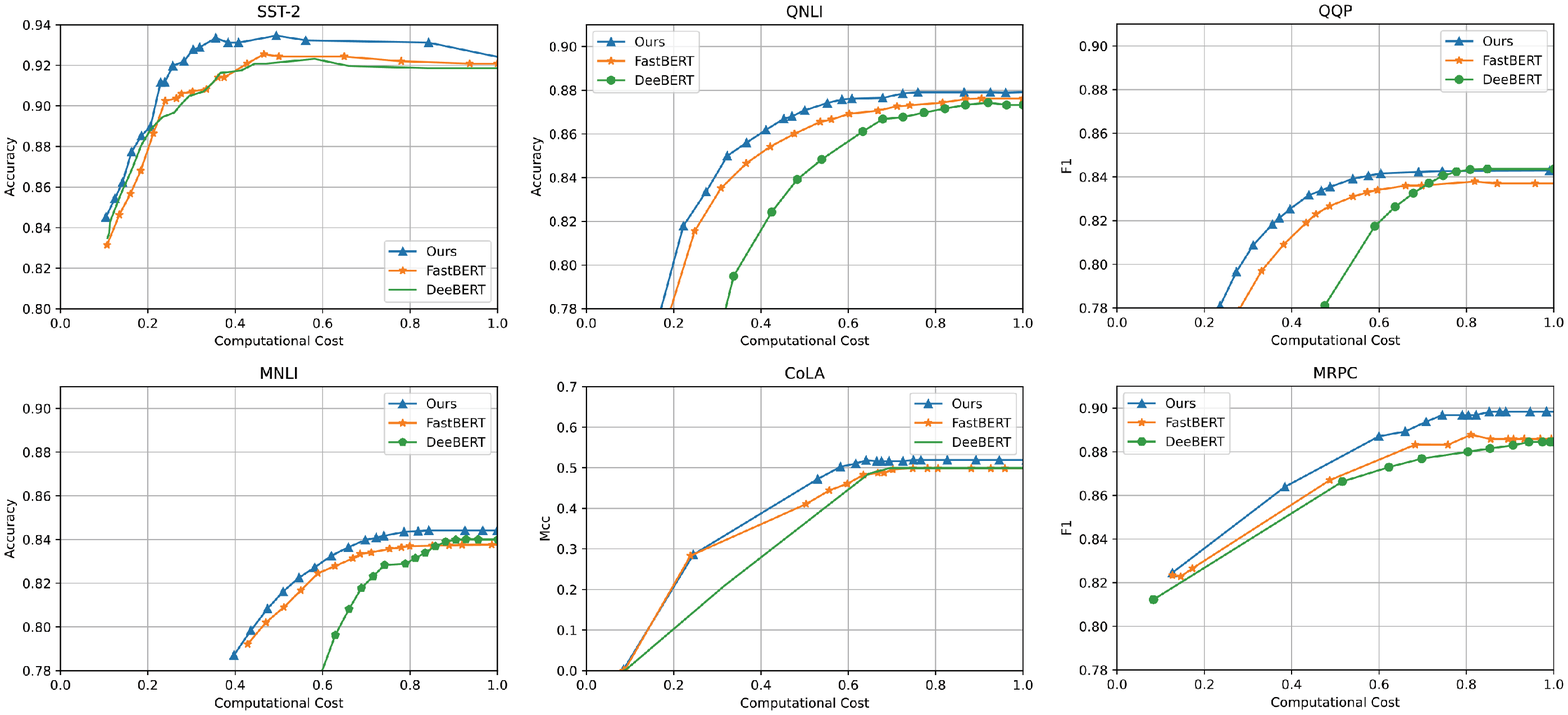}
    \caption{Comparison between baselines and \our \space on SST-2, QNLI, QQP, MNLI, CoLA, MRPC, showing the influence of computational cost on performance(Accuracy/F1 score/MCC).}
    \label{fig:comparsion}
\end{figure*}

\section{Experiments}

\subsection{Baselines}
We compare our method with three baselines:
\begin{itemize}
    \item \textbf{BERT.} A large-scale pre-trained language model based on Transformer, which is used as the backbone of our methods. In experiments, we only use the BERT-base model pre-trained by Google. \cite{devlin-etal-2019-bert}
    
    \item \textbf{DistillBERT.} It is a smaller transformer-based model by distilling the BERT. \cite{sanh2019distilbert}
    
    \item \textbf{Early exiting model.} The dynamic early exiting method for BERT is an effective method to accelerate BERT inference. We choose some classical methods like \textbf{DeeBERT} \cite{xin2020deebert} and \textbf{FastBERT} \cite{liu2020fastbert} as our baselines, which use the early exiting mechanism based on entropy.

\end{itemize}

\subsection{Datasets}
To verify the effectiveness of our methods, We conduct experiments on eight classification datasets of the GLUE benchmark\cite{wang2018glue}, including SST-2 \cite{socher2013recursive}, CoLA \cite{warstadt2019neural}, MRPC \cite{dolan2005automatically}, MNLI \cite{williams2017broad}, QQP \cite{chen2018quora}, QNLI \cite{rajpurkar2016squad}, RTE \cite{bentivogli2009fifth}, and WNLI \cite{levesque2012winograd}.

\subsection{Experimental Setup}
The experiments are done on an NVIDIA 2080Ti GPU. We adopt the same parameters for BERT, DeeBERT, FastBERT, and \our. In experiment, these models use the pre-trained parameters(\textbf{bert-base-uncased}) released by the Hugging Face Transformer Library \cite{wolf2019huggingface}. In the pre-trained parameters, the number of transformer layers, the dimension of hidden states, and the max length of the input sentence are set to 12, 768, and 128. We use AdamW \cite{kingma2014adam} to train these models with a default batch size of 32. For each task, we select the best fine-tuning learning rate(among 1e-5,2e-5,5e-5).

In the first stage of \our, we train the model with 5 epochs and select one with the best accuracy for the second stage. In the second stage, we adopt cross-layer contrastive learning to train the classifiers and train these classifiers for 4 epochs. We slightly tune the hyper-parameters across the different tasks.

Following prior work, our batch size of inference is set to 1 in the inference phase. 

In inference, We followed FastBERT and used FLOPs as an indicator to evaluate the computational cost. Generally speaking, the size of the FLOPs reflects the inference speed of the model, and the smaller the FLOPs of the model are, the shorter the inference time will be. Table \ref{tab:table1} presents the computational cost of each operation within the \our, which shows that the computational cost of the Skipping Gate and Classifier is much lower than the Transformer Layer.

\begin{table}[b]
\centering
\begin{tabular}{clllc}
\toprule
\multicolumn{2}{c}{Operation}              &&& FLOPs  \\ 
\midrule
\multicolumn{2}{c}{Each Transformer Layer} &&& 1811.8M \\
\midrule
\multicolumn{2}{c}{Each Classifier}        &&& 37.6M  \\
\midrule
\multicolumn{2}{c}{Each Skipping Gate}     &&& 37.4M \\
\bottomrule
\end{tabular}
\caption{This experiment evaluates each operation in the \our, using FLOPs as the indicator.}
\label{tab:table1}
\end{table}

\subsection{Main Results}
We evaluate these models in eight classification datasets of the GLUE benchmark and select different entropy thresholds to test the performance and sample-averaged FLOPs for DeeBERT, FastBERT and \our. We set three different entropy thresholds for early exiting models and compare the results with other baselines in Table \ref{tab:tabel2}. The results of the experiment show that our model achieves 2-3× computation reduction with minimal accuracy drops and even has a better performance than BERT. In the results, due to DistillBert cannot adaptively change the model architecture according to the sample complexity, it has a fixed computational complexity. In other words, the computational cost of DistillBert is independent of sample difficulty. In fact, the difficulty of each dataset is different, and in more datasets, our method has a better performance than DistillBert.

In Figure \ref{fig:comparsion}, we set several entropy thresholds and compare different models' tradeoffs in accuracy and computational cost on some GLUE datasets. From Figure \ref{fig:comparsion}, we conclude that: \begin{enumerate}
    \item  Compared with FastBERT and DeeBERT, \our\space have a better performance at the same computational cost, which proves our model is more effective than other approaches.
    \item In all datasets, the scores(accuracy/F1 score/FLOPs) of these models start to decline when the computational costs reach certain values which verify the redundant computation in original BERT, and \our\space dominates the performance under different computation costs comparing with FastBERT and DeeBERT.
\end{enumerate}

\subsection{Effectiveness of Skipping Mechanism}
The early exiting mechanism is based on the entropy of the corresponding classifier output. However, previous methods hardly reduce computation cost when samples are complex and difficult enough. Because the entropy of classifier output will be extremely high that leads to most exiting classifiers do not exit early. In Table \ref{tab:tabel2}, we can see that even if the threshold \emph{S} is set as 0.5, the DeeBERT and FastBERT still have an extremely high computational cost for RTE and WNLI dataset, which means early exiting mechanism is invalid on such situations. But \our \space achieves a relatively large computation reduction with minimal accuracy drop, which shows the skipping mechanism is effective. 

To further prove the effectiveness of the layer skipping mechanism, we disable the early exiting mechanism and only use the layer skipping mechanism during the inference phase. The empirical results are shown in Table \ref{tab:table3}, which proves that the skipping mechanism is able to reduce the computational cost compared with BERT. Moreover, combined Table \ref{tab:tabel2} and Table \ref{tab:table3}, FLOPs and Accuracy are hardly changed on the RTE and WNLI datasets, which further proves that the early exiting mechanism is invalid in difficult datasets and layer skipping mechanism plays an important role in our method on such situations.


\begin{table}[h]
\scalebox{0.9}{
\begin{tabular}{c|cc|cc|cc}
\toprule
\makecell{Dataset/\\ Model} & \multicolumn{2}{c|}{SST-2}    & \multicolumn{2}{c|}{RTE}      & \multicolumn{2}{c}{WNLI}                                     \\
\midrule                                                                          
& Acc  & \makecell{FLOPs\\ (cost)}
& Acc  & \makecell{FLOPs\\ (cost)}   
& Acc  & \makecell{FLOPs\\ (cost)}   \\
\midrule 
BERT                                                                      
& 93.2 & \makecell{21744\\ (100\%)} 
& 61.0 & \makecell{21744\\ (100\%)} 
& 56.3 & \makecell{21744\\ (100\%)} \\
\midrule
Skip                                                                      
& 93.1 & \makecell{14500\\ (66\%)}   
& 60.6 & \makecell{12190\\ (56\%)}   
& 56.3 & \makecell{6360\\(29\%)}\\  
\bottomrule
\end{tabular}}
\caption{\emph{Skip} represents our methods only use the skipping mechanism and disable the early exiting mechanism. This experiment is conducted on the SST-2, RTE and WNLI datasets.}
\label{tab:table3}
\end{table}

\subsection{Effectiveness of The Cross-Layer Contrastive Learning}
\begin{figure}[htbp]
    \centering
    \includegraphics[scale = 0.30]{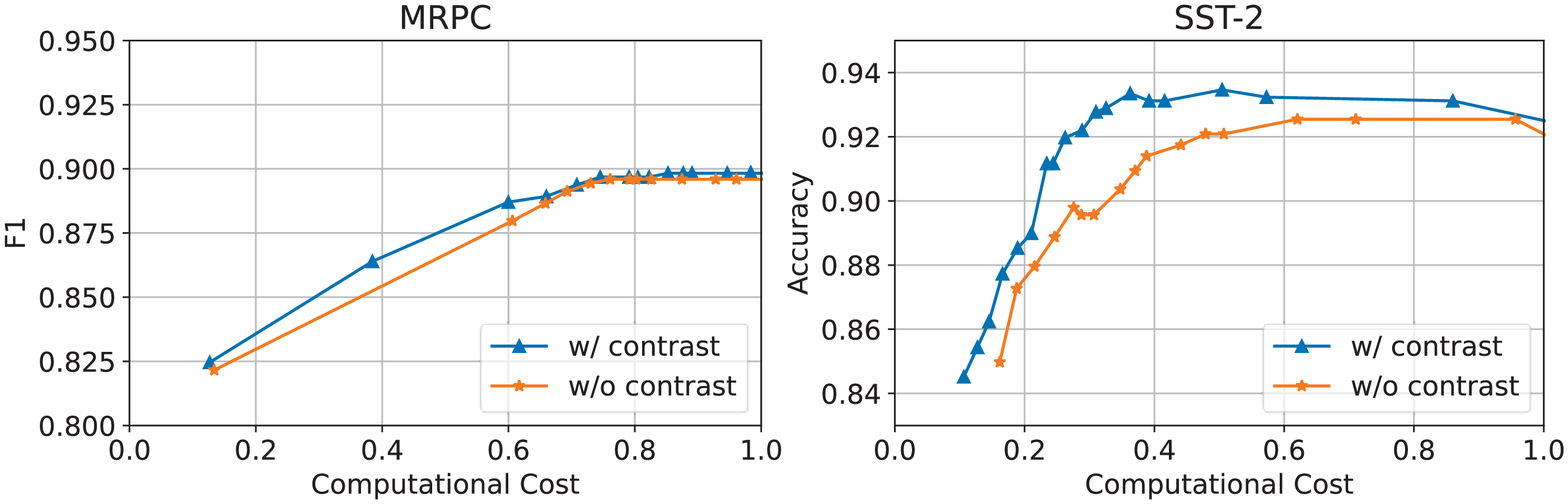}
    \caption{\emph{contrast} represents using the cross-layer contrastive learning, \emph{w/o contrast} represents not using the cross-layer contrastive learning.}
    \label{fig:contrast}
\end{figure}

We have introduced the cross-layer contrastive learning, which takes the same tokens across consecutive layers as positive pairs and the different tokens across consecutive layers as negative pairs. To further evaluate the effectiveness of contrastive learning, we compare the two cases of using contrastive learning and not using contrastive learning. Empirical results are shown in Figure \ref{fig:contrast}, which shows cross-layer contrastive learning is effective for further reducing the computation.

\subsection{Effectiveness of The Special Training Way}
In the previous section, we introduce the soft weight mechanism and hard weight mechanism. Meanwhile, we propose a special training way from soft weight mechanism to hard weight mechanism. In the experiment, we compare three different training strategies: (1) soft weight mechanism: using the continuous probability value of skipping gates for training but using the discrete value of skipping gates for inference, (2) hard weight mechanism: using the discrete value of skipping gates for training and inference and (3) the special training way: firstly using continuous probability value for training, and then using the discrete value for training. 

\begin{figure}[htbp]
    \centering
    \includegraphics[scale=0.30]{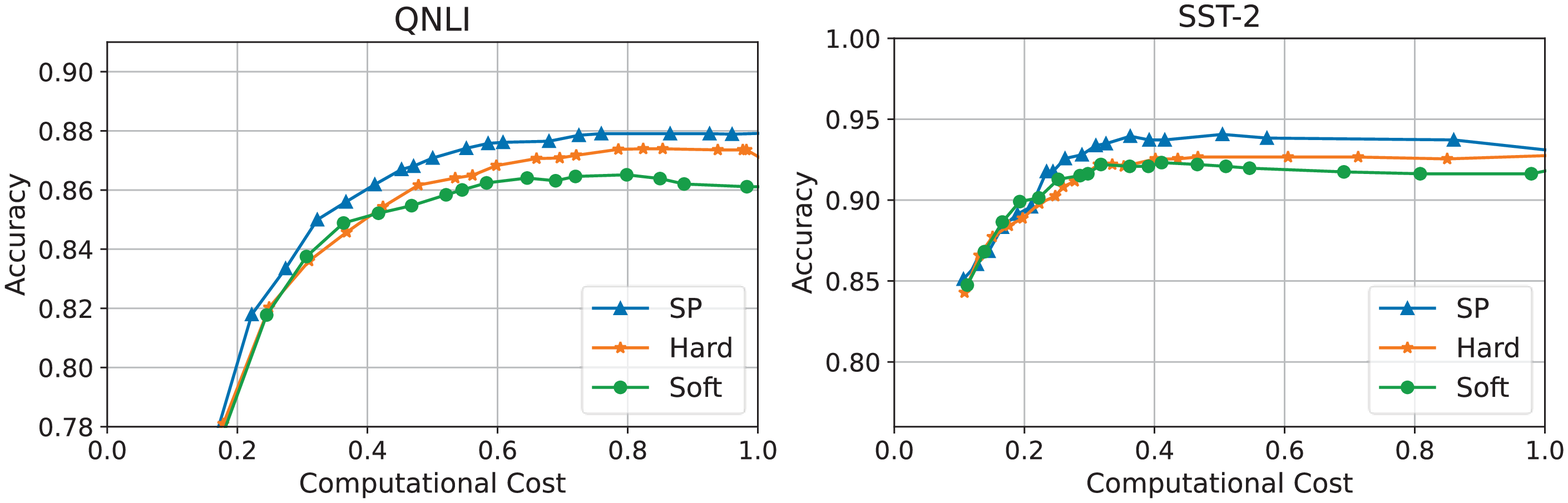}
    \caption{\emph{Hard} represents the hard weight mechanism, \emph{Soft} represents the soft weight mechanism, and \emph{SP} uses the special training way from soft mechanism to hard mechanism.}
    \label{fig:gate}
\end{figure}

We compare these methods on the SST-2 and QNLI datasets, and the results are shown in Figure \ref{fig:gate}. The results prove that the special training way is effective for balancing the inconsistent usage of skipping gates between training and inference phases.

\section{Conclusion}
In this paper, we propose a novel dynamic early exiting combined with layer skipping for BERT inference. To address the inconsistent usage of skipping gates in the inference and training phase, we propose hard weight mechanism and a special training way. In addition, we propose cross-layer contrastive learning and combine it into our training phases to boost the intermediate layers and classifiers which can further reduce computation cost. We evaluated our model on eight classification datasets of the GLUE benchmark. Empirical results show that \our \space could achieve performance comparable to BERT while significantly reducing computational cost. Compared to other dynamic early exiting models, \our \space obtain better accuracy with lower computation. Moreover, we conduct a series of ablation studies to demonstrate that each component is beneficial. In the future, we will adopt our methods to other pre-trained language models. We also will further study the combination of other acceleration methods and layer skipping mechanism.
\section*{Acknowledgements}
We would like to thank anonymous reviewers for their valuable comments and suggestions. This work has been supported in part by the NSFC (No. 62272411), the Zhejiang NSF (LR21F020004), Ant Group and Alibaba-Zhejiang University Joint Research Institute of Frontier Technologies.

\bibliographystyle{named}
\bibliography{ijcai23}
\newpage
~
\newpage
\subsection*{A. Implementation Details}
\subsubsection{Algorithm}
The inference processes of \our~ can be described as Algorithm~\ref{algorithm1}.
\begin{algorithm}
	\renewcommand{\algorithmicrequire}{\textbf{Require:}}
	\caption{Inference ($Input: \mathbf{X}_0$)}
	\label{algorithm1}
	\begin{algorithmic}[1]
	\FOR{$i=1$ to $L$}
        \IF{$\mathcal{G}^i(\mathbf{X}^{i-1}) \geq 0.5$} 
        \STATE $\mathbf{\mathbf{X}^{i}} = \mathbf{\mathbf{X}^{i-1}}$ \\ 
        \IF{i $\neq$ L}
        \STATE{continue} 
        \ENDIF
        \ELSE
        \STATE $\mathbf{X}^{i} = \mathcal{E}^i(\mathbf{X}^{i-1})$
        \ENDIF \\
        $\mathbf{z}^i = \mathcal{C}^i(\mathbf{X}^i)$
        \IF{$\operatorname{Entropy}(\mathbf{z}^i)<S$}
        \STATE{return $\mathbf{z}^i$ }
        \ENDIF 
    \ENDFOR \\
    return $\mathbf{z}^L$
	\end{algorithmic}  
\end{algorithm}

\subsection*{B. The Impact of Different Entropy Thresholds}
\begin{figure}[htb]
    \centering
    \includegraphics[scale = 0.4]{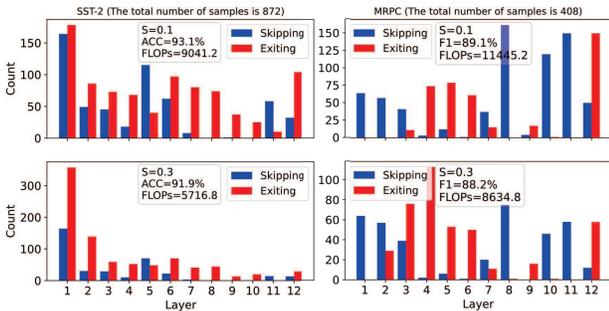}
    \vspace{-1em}
    \caption{Statistics about how often the model uses skipping and early exiting in different entropy threshold \emph{S}.}
    \vspace{-1em}
    \label{fig:frequence}
\end{figure}

We counted the number of skipping and exiting on each layer in different entropy thresholds \emph{S}. From the Fig \ref{fig:frequence}, We intuitively see that the early exiting classifiers and skipping gates can play their role in the inference stage.

\subsection*{C. Motivation of Cross-Layer Contrastive Learning}
\begin{figure}[htb]
    \centering
    \includegraphics[scale = 0.4]{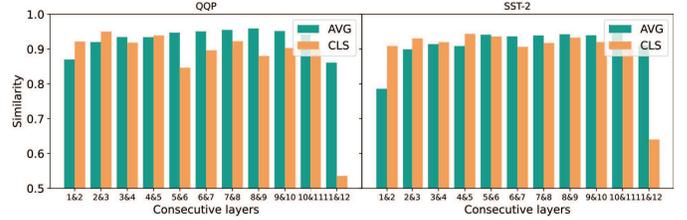}
    \vspace{-1em}
    \caption{Token representations' similarity of two consecutive layers. CLS represents similarity of [CLS] token representations in the two consecutive layers. AVG represents the average similarity of all tokens(except [CLS]).}
    \vspace{-1em}
    \label{fig:similarity}
\end{figure}

In cross-layer contrastive learning(CCL), we assume that each token should have similar semantics across the consecutive layers because their representations do not change drastically. Empirically, we find the representations of all tokens change slightly in the consecutive layers(i.e., the cosine similarity of the same token in the consecutive layer is over 0.9) according to Fig~\ref{fig:similarity}. This phenomenon verifies our motivation. As for [CLS] token which only changes drastically in the last layer, we will not use this layer to do CCL.

\subsection*{D. Ablation Study on More Datasets}
\begin{figure}[tbp]
    \centering
    \includegraphics[scale = 0.60]{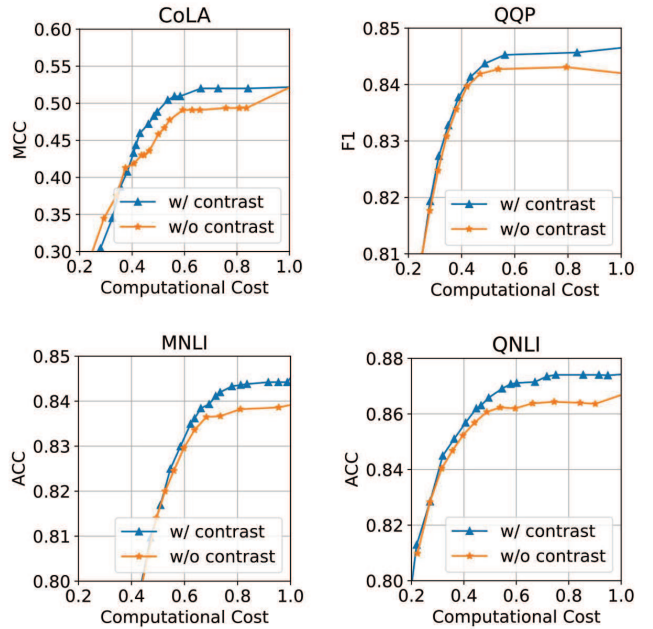}
    \vspace{-0.8em}
    \caption{Ablation study of CCL on other datasets.}
    \vspace{-0.8em}
    \label{fig:supplement}
\end{figure}

Table \ref{tab:table3} and Figure \ref{fig:contrast} only show ablation study on partial datasets due to the limited space. There, we provide more experiments in Table \ref{tab:table4} and Figure \ref{fig:supplement}. 

\begin{table}[htb]
\centering
{
\begin{tabular}{c|cc|cc|cc}
\hline
\makecell{Dataset/\\ Model} & \multicolumn{2}{c|}{MRPC}                                      & \multicolumn{2}{c|}{CoLA}   
& \multicolumn{2}{c}{QNLI}\\
\hline                                                                          
& F1  & \makecell{FLOPs\\ (cost)}
& MCC  & \makecell{FLOPs\\ (cost)}
& ACC  & \makecell{FLOPs\\ (cost)}  
   \\
\hline 
BERT                                                                      
& 88.5 & \makecell{21744\\ (100\%)} 
& 50.8 & \makecell{21744\\ (100\%)}
& 88.1 & \makecell{21744\\ (100\%)}
 \\
\hline
Skip                                                                      
& 88.4 & \makecell{14894\\ (68\%)}   
& 50.1 & \makecell{16400\\ (74\%)}
& 87.3 & \makecell{16568\\ (74\%)}
\\  
\hline
\end{tabular}}

\caption{Ablation study of skipping mechanism on other datasets.}

\label{tab:table4}
\end{table}

\end{document}